\documentclass[conference]{IEEEtran}
\IEEEoverridecommandlockouts
\usepackage{cite}
\usepackage{amsmath,amssymb,amsfonts}
\usepackage{graphicx}
\usepackage{textcomp}
\usepackage{xcolor}
\def\BibTeX{{\rm B\kern-.05em{\sc i\kern-.025em b}\kern-.08em
    T\kern-.1667em\lower.7ex\hbox{E}\kern-.125emX}}

\usepackage{algorithm, amsmath, amsfonts, amsthm, mleftright}
\usepackage{algcompatible}

\theoremstyle{definition}
\newtheorem{defi}{Def.\null}

\newcommand{\1}{\mbox{1}\hspace{-0.25em}\mbox{l}}

\usepackage{algpseudocode}
\usepackage{comment}

\usepackage{microtype}

\title{Towards Interpretable and Reliable Reading Comprehension: A Pipeline Model with Unanswerability Prediction}

\author{\IEEEauthorblockN{1\textsuperscript{st} Kosuke Nishida}
\IEEEauthorblockA{\textit{NTT Media Intelligence Laboratories} \\
\textit{NTT Corporation}\\
Yokosuka, Kanagawa, Japan \\
kosuke.nishida.ap@hco.ntt.co.jp}
\and
\IEEEauthorblockN{2\textsuperscript{nd} Kyosuke Nishida}
\IEEEauthorblockA{\textit{NTT Media Intelligence Laboratories} \\
\textit{NTT Corporation}\\
Yokosuka, Kanagawa, Japan \\
kyosuke.nishida.rx@hco.ntt.co.jp}
\and
\IEEEauthorblockN{3\textsuperscript{rd} Itsumi Saito}
\IEEEauthorblockA{\textit{NTT Media Intelligence Laboratories} \\
\textit{NTT Corporation}\\
Yokosuka, Kanagawa, Japan \\
itumi.saito.df@hco.ntt.co.jp}
\and
\IEEEauthorblockN{4\textsuperscript{th} Sen Yoshida}
\IEEEauthorblockA{\textit{NTT Media Intelligence Laboratories} \\
\textit{NTT Corporation}\\
Yokosuka, Kanagawa, Japan \\
sen.yoshida.tu@hco.ntt.co.jp}
}

\date{}

\begin{document}
\maketitle
\begin{abstract}
Multi-hop QA with annotated supporting facts, which is the task of reading comprehension (RC) considering the interpretability of the answer, has been extensively studied. In this study, we define an \textit{interpretable reading comprehension} (IRC) model as a pipeline model  with the capability of predicting unanswerable queries. The IRC model justifies the answer prediction by establishing consistency between the predicted supporting facts and the actual rationale for interpretability. The IRC model detects unanswerable questions, instead of outputting the answer forcibly based on the insufficient information, to ensure the reliability of the answer. We also propose an end-to-end training method for the pipeline RC model. To evaluate the interpretability and the reliability, we conducted the experiments considering unanswerability in a multi-hop question for a given passage. We show that our end-to-end trainable pipeline model outperformed a non-interpretable model on our modified HotpotQA dataset. Experimental results also show that the IRC model achieves comparable results to the previous non-interpretable models in spite of the trade-off between prediction performance and interpretability.
\end{abstract}

\begin{IEEEkeywords}
interpretability, reading comprehension, question answering
\end{IEEEkeywords}

\section{Introduction}
There is increasing demand for automated decision-making by using artificial intelligence (AI) \cite{xai1,xai2}. 
Moreover, reading comprehension (RC), a task to answer a question with textual sources, is an important topic in AI research. 
It is tackled with deep neural models such as BERT \cite{bert}. However, it is difficult for a neural network black box to provide reasons for its predictions.
This problem affects the perceived reliability and interpretability of RC models used in social contexts.

\begin{figure}
    \begin{center}
		\includegraphics[width =75mm]{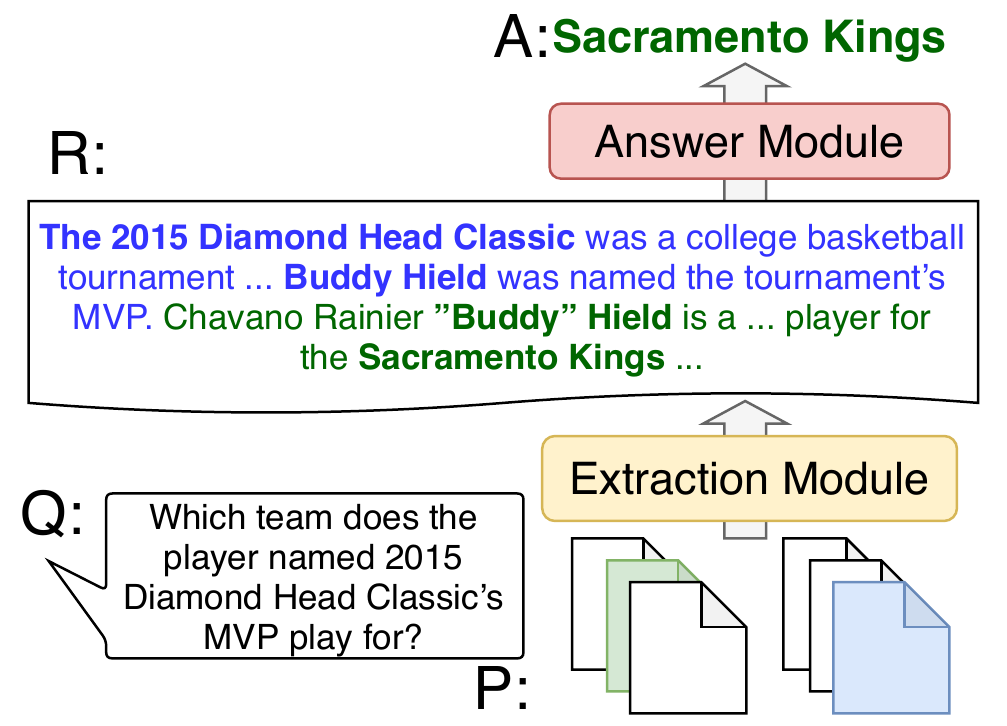}
		\caption{Concept of interpretable reading comprehension. The QA module predicts "Can Not Answer" if the output of the extraction module does not include both ground-truth SFs.}
		\label{fig:concept}
	\end{center}
\end{figure}

Multi-hop QA datasets with annotated supporting facts (SFs) \cite{hotpot, multirc} have been proposed for the RC task in order to develop a way to interpret the model’s predictions. Fig.~\ref{fig:concept} shows an example in the HotpotQA dataset. In HotpotQA, the model outputs an answer $A$ and SFs $R$ in response to a query $Q$ on a given passage $P$. The SFs are a set of sentences that describe the reasoning behind the answer. 

Our goal is to enable the model to predict the answer and give a corresponding rationale for the sake of \textit{interpretablity} and \textit{reliablity} in RC. First, to increase interpretability, the model should extract rationale sentences from the passage that are truly required for reasoning; these sentences are called SFs in HotpotQA. The previous models can predict SFs independently of the predicted answer~\cite{qfe, sae}. Therefore, the models can predict sentences that are not actually used for reasoning in the models as SFs. In this study, we propose to use a pipeline model for RC that first extracts rationales from the passage and then predicts the answer; this model has been used in the past for interpretation of text classification tasks \cite{rational_lei2016, rational_bastings2019, rationale_yu2019}.  
The proposed interpretable reading comprehension (IRC) model can justify the model prediction by ensuring consistency between the predicted SFs and the actual rationale used by the RC module.
Fig.~\ref{fig:concept} shows the concept of IRC.


Next, to increase the reliability of the justified answer prediction, we consider that the model should answer with valid reasoning based on sufficient information.
It is known that HotpotQA has a reasoning shortcut problem~\cite{shortcut1}. 
This is a phenomenon in which the model can directly locate the answer from the only one sentence (the green sentence in the example), despite the fact that there are multiple SFs (both statements).
Since the HotpotQA dataset will always have an answer for each query, 
the model learns to answer with dishonest reasoning.

We prohibit the model from answering with the invalid reasoning by training it to detect unanswerable queries. We introduce a `Can Not Answer' (CNA) label as an answer candidate. The model can avoid answering with invalid reasoning by outputting a CNA label.

We analyzed the reliability and the interpretability of RC models from two points of view. Firstly, the unanswerability detection has the effect of preventing reasoning shortcuts. 
To show this, we made simple modifications to the evaluation setting of HotpotQA. In this setting, the model must output a CNA label if the passage does not have sufficient information. Secondly, the end-to-end learning of the pipeline model with the CNA label enables the predicted rationale to explain the valid reasoning behind the answer. To show how the answer module finds an answer and determines that the reasoning is valid, we conducted a qualitative analysis of the rationale prediction.

Our main contributions are as follows.
\begin{itemize}
    \item 
    We define 
    IRC as an end-to-end trainable pipeline model with a `Can Not Answer' prediction.
    For interpretability, the model justifies its prediction by ensuring consistency between the predicted SFs and the actual rationale. For reliability, the model can detect unanswerable queries instead of outputting an answer forcibly with invalid reasoning.
    \item We conducted experiments on the modified HotpotQA dataset that contains a CNA label as an answer candidate to evaluate the model's reliability.
    We show that the IRC model outperforms a non-interpretable model.
    \item We show that the IRC model achieves comparable results to the previous non-interpretable models in the original setting of HotpotQA. Although there is a trade-off between prediction performance and interpretability \cite{xai2}, the proposed IRC model maintains prediction performance while increasing interpretability.
    \item Through a qualitative analysis, we discuss how the answer module locates an answer and determines that the reasoning is valid.
    Even if the predicted rationales include non-gold SFs, they often play a role for reasoning in the model, such as by being supplementary explanations of entities.
\end{itemize}

\section{Preliminaries}
\subsection{Task Definition}
We define the interpretable reading comprehension to provide consistency to the predicted SFs and rationales behind the model’s prediction.
\begin{defi}[Interpretable Reading Comprehension] 
    We say that an RC model is interpretable if it has two modules with the following inputs and outputs.
    \begin{itemize}
        \item \textbf{Extraction Module:} The inputs are a passage $P$ and query $Q$. The output is the rationale $R$.
        \item \textbf{Answer Module:} The inputs are a rationale $R$ and query $Q$. The output is the answer $A$.
    \end{itemize}

    The rationale
    is a set of sentences. The answer is the label or the span in the passage. The candidate answers include the CNA label, which represents that the passage is insufficient as a reason. 
    In what follows, we define $A^*$ to be the ground-truth answer and $\hat{A}$ to be the predicted answer.
\end{defi}

The IRC model can justify the model prediction. That is, because the answer module only uses the information in the rationale $R$, we can avoid a situation where the answer module implicitly depends on other information.
Here, \cite{eraser} divided the explainable models to hard and soft approaches. The IRC model is a hard approach, and this approach is faithful because of the discrete extraction of the rationale. In comparison, the soft approach may still use all sentences in the passage to predict particular answer independently on the SFs prediction. We call such model a one-stage model. 

\subsection{Related Work}
\subsubsection{Interpretable NLP}
One of the goals of explainable AI is to `produce more explainable models, while maintaining a high level of learning performance' \cite{xai2}. We contributed to this goal because the IRC model justifies the answer without affecting answer performance. 

There are various approaches to making interpretable models \cite{mythos}; we focused on hard selection of the input for the justification and avoidance of the invalid reasoning. 
The hard approach of the pipeline models has been used for text classification \cite{
rational_bastings2019, rationale_yu2019}. 
There is a limitation when it comes to applying the pipeline models used in text classification tasks to the RC tasks because the rationale of the text classification tasks is a few words such as positive words. 
\cite{rationale_lehman2019, rationale_chen2019} provided a new task and dataset for medical and fact-checking usages focusing on interpretability. They also used hard selection of the rationale with a pipeline model.


\subsubsection{Reading Comprehension}
Multi-hop QA was proposed in order to verify the ability to reason over multiple text \cite{qangaroo, hotpot}. Multi-hop reasoning is essential to evaluating the interpretability of RC, because in traditional RC datasets such as SQuAD \cite{squad}, the query can be answered with the single sentence that has the answer \cite{rc_suga}. SQuAD2.0 \cite{squad2} has the no-answer option, but reasoning on the basis of only one sentence is not suitable for an evaluation of interpretability.

We chose the HotpotQA dataset for our study, although there are a few multi-hop QA datasets with manually annotated SFs. In particular, MultiRC \cite{multirc} is a multiple-choice dataset, where, unlike HotpotQA, the number of correct answer options for each question is not pre-specified. In this dataset, the SFs (used sentences) for answering the question (choosing all correct answer options) are provided. However, the actual SFs required for each answer option are different, but are not annotated individually. A future challenge will be to extend our model to make it able to answer questions from several different perspectives and to explain these perspectives as its rationale.

We should mention the work using tpipeline models for RC. Some studies have used pipeline models for efficient computation \cite{coarse_to_fine, efficient_rc}. \cite{pipe_hotpot} proposed a pipeline model for multi-hop QA. Their pipeline model extracts the relevant sentences from the passage and then predicts the answer and the SFs from the sentences. They showed that the sentence-level extraction is a strong approach to multi-hop QA. However, their model does not have our interpretable structure, because their answer module relies on information other than the predicted SFs. Neither of these studies proposed any end-to-end training method.

\cite{shortcut1} pointed out the reasoning shortcut and created an adversarial dataset by generating fake answers. The reasoning shortcut was tackled by decomposing the query to sub-queries \cite{subquestion1, subquestion2, subquestion3}. They used complex models to combine the single-hop sub-queries. By comparison, the IRC model is characterized by the detection of CNA label and justification of the prediction.

\section{Proposed Method}
\subsection{Model Architecture}
\label{ssec:model}
\begin{figure}
    \begin{center}
		\includegraphics[width =75mm]{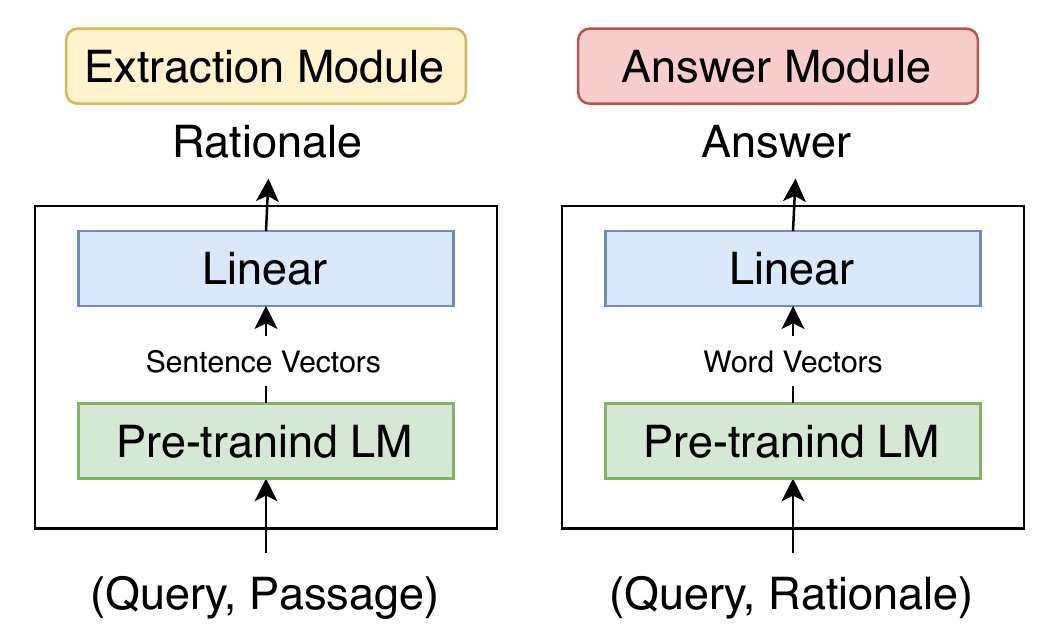}
		\caption{Proposed IRC model.}
		\label{fig:model}
	\end{center}
\end{figure}

Our pipeline model consists of an extraction module and answer module. Each module has a language understanding layer and a task-specific linear layer. The language understanding layer is a pre-trained language model (LM), and we used BERT$_\textrm{base}$. 
Fig.~\ref{fig:model} shows the model architecture.

\textbf{Extraction Module:}
We input the token sequence 
 [`[CLS$^Q$]'; query; `[SEP$^Q$]'; `[CLS$^S$]';  sentence~1; `[SEP$^S$]'; $\cdots$; `[CLS$^S$]'; sentence~$N^s$;  `[SEP$^S$]'] to BERT, 
 where `[CLS$^Q$]',`[SEP$^Q$]',`[CLS$^S$]', and `[SEP$^S$]' are special tokens and 
 $N^s$ is the number of sentences in the passage. The $i$-th `[CLS$^S$]' token output is the $i$-th sentence representation $s_i \in \mathbb{R}^d$, where $d$ is the embedding dimension. 
 We obtain the $i$-th sentence score from the linear layer output: 
 \begin{align}
 p_i =\textrm{sigmoid} (W^s s_i +b^s),
 \end{align}
where $W^s \in \mathbb{R}^{d}, b^s \in \mathbb{R}$ are trainable parameters. 

\textbf{Answer Module:}
We input the token sequence [`[CLS]'; query ; `[SEP]'; rationale; `[SEP]'] to BERT,
where `[CLS]' and `[SEP]' are special tokens.
The output sequence is denoted as $H^a$.
The answer layer has two linear transformations
\begin{align}
    \label{eq:label}
 c &= W^c h^a_0  + b^c \in \mathbb{R}^{N^c},\\
  \label{eq:answer} 
 [a^s_i; a^e_i]^\top &= W^a h^a_i + b^a \in \mathbb{R}^{2}.
 \end{align}

Equation \eqref{eq:label} is the answer label classification. The number of answer labels $N^c$ depends on the task.  The candidates of the answer labels include `Span' and CNA. The `Span' label means that the query should be answered by span extraction. 
Equation \eqref{eq:answer} is the answer span extraction.
Each dimension is the score at the start or end of the answer span.
$W^c \in \mathbb{R}^{N^c \times d}, b^c \in \mathbb{R}^{N^c}$, $W^a \in \mathbb{R}^{2 \times d}, b^a \in \mathbb{R}^2$ are trainable parameters.

\subsection{Inference}
In order to extract the actually required sentences for the reasoning, we use the CNA classification score to determine whether the extracted sentences are sufficient for reasoning.
Firstly, the extraction module outputs the sentences with $p_i > \alpha$ as a rationale $\hat{R}$.
Then, if the answer module predicts CNA, we add sentences determined by
$\textrm{argmax}_{i  \not \in \hat{R}}p_i$ 
to the rationale $\hat{R}$. This operation continues until a stop criterion, the max number of rationales $N^r$, is met. 
$N^r, \alpha$ are hyperparameters. 

\subsection{Training}
\textbf{Loss Function:}
The loss of the extraction module $L^{\textrm{R}}$ is the binary cross entropy loss between the sentence extraction probabilities $\{p_i\}$ and the ground-truth rationale $R^* \in 2^{N^s}$.
The loss of the answer module $L^{\textrm{A}}$ is the sum of the cross entropy losses of the answer label classification, the start token of the answer span, and the end token.


\textbf{Pre-Training:}
We pre-train the extraction module and the answer module separately. The extraction module is trained with $L^{\textrm{R}}$, and the answer module is trained with $L^{\textrm{A}}$. The input for the answer module is the ground-truth rationale $R^*$.

\textbf{End-to-End Training:}
We propose an end-to-end training algorithm for learning the interactions between the two modules. Here, 
we consider three different losses. 
The end-to-end answer loss $L^{\textrm{E2E}}$ is the loss to locate the answer. 
The rational extraction loss $L^{\textrm{R}}$ is the same as in the pre-training.
The no-answer penalty $L^{\textrm{NA}}$ helps the extraction module not to miss the sentence with the answer span.
 
Beforehand, the ground-truth answer is replaced by a CNA label if the result of the sampling $\hat{R}$ does not include the whole ground-truth rationale $R^*$. This is because that the answer module must learn how to determine if the query cannot be answered the sentences extracted from the passage. We also augment one CNA sample for each query by performing negative sampling on the passage.
 
First, we explain the end-to-end answer loss, $L^{\textrm{E2E}}$.
Ideally, we want to apply the loss on the probability
\begin{align}
\Pr(A^*|P,Q) = \sum_{R \in 2^{N^s}} \Pr(A^*|R,Q)\Pr(R|P,Q)
\end{align}
to $L^{\textrm{E2E}}$. However, for computational reasons, we use the predicted rationale $\hat{R}$ to calculate $L^{\textrm{E2E}}$.

The operation of extracting the rationale $\hat{R}$ is not differentiable, so the gradient obtained in the answer module does not backpropagate to the extraction module. Therefore, we use a straight-through Gumbel-softmax estimator \cite{gs1, gs_bengio}. 
The sentences are extracted by sampling in accordance with a discrete distribution.
Let $g_i, g'_i ~(i =1, \cdots, N^s)$ be i.i.d.~samples from the Gumbel distribution\footnote{$g =-\log (-\log (u)), \quad u \sim \textrm{Uniform}(0,1)$}.
We sample a set of sentences by using the Gumbel-softmax trick \cite{gumbel,gs_trick}.
The $i$-th sentence is extracted if $g_i+\log p_i  > g'_i + \log (1-p_i)$.
The continuous relaxation is
\begin{align}
z_i =\dfrac{\exp (\dfrac{g_i + \log p_i}{\tau} )}
    {\exp (\dfrac{g_i + \log p_i}{\tau} ) + \exp (\dfrac{g'_i + \log (1-p_i)}{\tau} )},
\end{align}
where $\tau$ is the hyperparameter of the temperature.
Let $\1_{\hat{R}}(i)$ be the indicator function that returns 1 if the $i$-th sentence is extracted.
On the backward path, we use the straight-through Gumbel-softmax estimator $\nabla \1_{\hat{R}}(i) \approx \nabla z_i$ as the approximation.

By Jensen's inequality, the loss of the answer module $L^{\textrm{A}}$ 
is an upper bound of intractable
$L^{\textrm{E2E}}$; 
\begin{align}
\begin{split}
    L^{\textrm{E2E}} &= - \log \Pr(A^*|P,Q) \\
    &= -\log \sum_{R} \Pr(A^*|R,Q)\Pr(R|P,Q)\\
    &\leq - \sum_{R} \Pr(R|P,Q) \log \Pr(A^*|R,Q)\\
    &= -\textrm{E}_{R\sim \Pr(R|P,Q)}\left[ \log \Pr(A^*|R,Q) \right]\\
    &\approx -\log \Pr(A^*|\hat{R},Q) =L^{\textrm{A}}.
\end{split}
\end{align}
The last approximation uses the Gumbel-softmax trick. Therefore, we can learn the model in the end-to-end fashion with $L^{\textrm{A}}$ instead of $L^{\textrm{E2E}}$.

Then, we introduce a no answer penalty $L^{\textrm{NA}}$.
Because the extracted sentences are sampled independently of the ground-truth rationale $R^*$, we emphasize the sentences including the answer. The penalty is defined as 
\begin{align}
L^{\textrm{NA}} = \max(0, \max_{\hat{r} \in \hat{R}} W^s s_{\hat{r}} - \max_{i \in S_A} W^s s_i),
\end{align}
where $S_A$ is the set of sentences with the ground-truth answer span. $L^{\textrm{NA}}$ is zero if one of the sentences with the ground-truth answer has a higher score than that of any of the extracted sentences.

As a result, the loss function is 
$L^{\textrm{A}} + \lambda^{\textrm{R}} L^{\textrm{R}} +  \lambda^{\textrm{NA}} L^{\textrm{NA}}$,
where $\lambda^{\cdot}$ are hyperparameters. 
$L^{\textrm{R}}$, and $L^{\textrm{NA}}$ help to stabilize the training which is otherwise affected by the approximation.

\section{Evaluation}
\subsection{Dataset and Metrics}
We used HotpotQA, which is a multi-hop QA dataset with manually annotated SFs consisting of multiple sentences. In HotpotQA, the query $Q$ refers to the content of two paragraphs from two Wikipedia articles. Each passage $P$ consists of ten paragraphs. HotpotQA has two settings. In the distractor setting, the passage has two gold paragraphs, and the other eight paragraphs are selected in accordance with the TF-IDF similarity scores. 
The fullwiki setting is the RC task including the retrievals from a preprocessed Wikipedia dump. 
This setting also provides ten paragraphs that are retrieved in accordance with the TF-IDF similarity for convenience. The outputs are the answer $A$ and the SFs, which are the rationale $R$ in our IRC definition. The ground-truth answer $A^*$ consists of the answer labels $\{\textrm{`Yes', `No', `Span'}\}$ and a span in the passage. The answer span exists only if the ground-truth answer label is `Span'. We added CNA to the answer labels. The rationale $R$ is a set of sentence IDs.

The answers were evaluated in terms of exact matching (EM) and partial matching (F1) on the string. The rationales were evaluated in terms of EM and F1 on the set of sentence IDs. The leaderboard were evaluated on the test set; the others were evaluated on the development set.

\subsection{Model Implementations}
\label{ssec:imple}

We used two extra modules to adapt our model to HotpotQA.

\textbf{Paragraph Ranker:}
To select the input of the extraction module from the whole passage consisting of $N^p$ paragraphs, we used the paragraph ranker based on the SAE paragraph ranker \cite{sae}.
The paragraph ranker aims to retrieve a paragraph pair including the content referred to by the query.
The input is the query and a paragraph.
The output is the score of the $i$-th paragraph. It was calculated as $S^{\textrm{SAE}}_i$,
and we ranked the paragraph pairs $\{(i, j)\}$ according to $S^{\textrm{SAE}}_i + S^{\textrm{SAE}}_j$.

To train the main model, we inputted gold paragraph pairs without the paragraph ranker. Moreover, we used negative sampling to make the CNA samples. Here, a randomly selected paragraph of the gold paragraph pairs was replaced with a non-gold paragraph including the sentence with the highest TF-IDF similarity to the query. 

\textbf{Answer Re-ranking:}
For inference, we used the paragraph ranker to extract the top-$K$ paragraph pairs, which were then used as the inputs of the extraction module. 
For each pair, 
we used the extraction module and the answer module as above.
Finally, we reranked the $K$ 
pairs according to 
\begin{align}
    \dfrac{1}{2}(S^{\textrm{SAE}}_i + S^{\textrm{SAE}}_j) - 
    \dfrac{\exp{( c_{\textrm{CNA}} )}}{\sum_l \exp{(c_l)}},
\label{eq:rerank}
\end{align}
where the answer label score $c$ is calculated using \eqref{eq:label}. 

Algorithm \ref{alg} 
shows the pseudo-code of the model for inference in HotpotQA.
\begin{algorithm*}[t!]
	\caption{Pipeline Model in Inference}
	\label{alg}
	\begin{algorithmic}[1]
		\Require Passage $P$, Query $Q$, Hyperparameter $K, \alpha, N^r$
		\State Retrieve the top-$K$ paragraph pairs with the paragraph ranker
		\For {Select the $k$-th paragraph pair from the top-$K$ paragraph pairs}
		\State Obtain the sentence scores $p_i$ with the extraction module from the $k$-th paragraph pair
		\State Select the sentences $i$ with $p_i > \alpha$ as rationale $\hat{R}$
		\State Select the answer $\hat{A}$ with the answer module from rationale $\hat{R}$
		\While {Answer $\hat{A}$ is CNA and $|\hat{R}| < N^{\textrm{r}}$ \do} 
		\State Add sentence $\textrm{argmax}_{i  \not \in \hat{R}} p_i$ to the rationale $\hat{R}$
        \State Select the answer $\hat{A}$ with the answer module from rationale $\hat{R}$
        \EndWhile
        \State Add the answer $\hat{A}$ and the rationale $\hat{R}$ of the $k$-th paragraph pair to the prediction candidates
        \EndFor
    \State Rerank the answer and the rationale in the $K$ prediction candidates with the answer reranking \eqref{eq:rerank}
    \State Output the answer and the rationale
	\end{algorithmic}
\end{algorithm*}

\subsection{Evaluated Models}
We evaluated the prediction of the IRC model and the one-stage baseline model.
The one-stage model simultaneously outputs the answer and the SFs with a shared module from the passage. Therefore, in comparison with the IRC model, the predicted SFs of the one-stage model can not justify the answer prediction. 

As in the IRC model, the input passage of the one-stage model was a paragraph pair retrieved with the paragraph ranker. The model predicted the answer and the SFs from the output representations of BERT and the linear layers. The answer prediction and the re-ranking of the paragraphs were the same as those in the IRC model. 
For training, the loss function was $L^{\textrm{R}}+L^{\textrm{A}}$. 

We trained the models on one NVIDIA Tesla P100 GPU (16GB). The training took less than one day with gradient accumulation. 
We used the PyTorch implementation of BERT\footnote{https://github.com/huggingface/pytorch-transformers}.

The hyperparameter settings are in Table \ref{tab:hyper}. 
The threshold values for the rationale extraction, $\alpha$, 
were determined from 0 to 0.9 by 0.1 to maximize the answer’s F1 score for the IRC model and to maximize the SFs’ F1 score for the one-stage model, because the answer performance of the one-stage model does not depend on the SFs’ prediction. 
The rest of the implementation including that of the optimizer followed the Pytorch BERT implementation.

\begin{table}[t]
\begin{center}
    	\caption{Hyperparameters.}
	    \begin{tabular}{cc}\hline
				 $\lambda^{\textrm{R}}$ in loss function & 0.1\\
				 $\lambda^{\textrm{NA}}$ in loss function & 1\\
				 max number of rationales $N^r$ & 5 \\
				 number of paragraph pairs $K$ & 3 \\
				 batch size & 72  \\
				 epochs in pre-training & 5 \\
				 epochs in end-to-end training & 2 \\
				 max sequence length & 512 \\
				 max sentence length & 160 \\
		         max number of sentences & 20 \\
				 max query length & 64 \\
		         temperature in Gumbel softmax $\tau$ & 0.5 \\
				 learning rate & 5e-5 \\
				 weight decay & 0 \\ \hline
				\end{tabular}
	\label{tab:hyper}
\end{center}
\end{table}

\subsection{Evaluation in HotpotQA with considering 'Can Not Answer'}
First, we evaluated the ability of the model to avoid answering with reasoning based on insufficient information. We conducted experiments in the fullwiki and CNA setting (`Fullwiki+CNA').

\begin{table}[t]
	\begin{center}
		\caption{
		Data statistics of the Fullwiki+CNA setting.}
			\begin{tabular}{c|c|c}\hline
				Can Not Answer & \# Absent Gold SFs & \# Examples \\\hline
                 & 0 & 2089\\
                 \checkmark & 1 & 3418\\
                 \checkmark & 2 & 1504\\
                 \checkmark & 3+ & 374 \\ \hline
		\end{tabular}
		\label{tab:stat}
	\end{center}
\end{table}

\subsubsection{Experimental Setup}
We used the ten paragraphs published as the fullwiki setting for the passage. 
However, a passage fully based on TF-IDF retrieval might not have the ground-truth answer or ground-truth SFs, and in such cases the model cannot provide any correct reasoning. 
Therefore, we replaced the ground-truth answer label with CNA if the passage did not have both gold paragraphs.
We removed the sentences not included in the passage from the ground-truth SFs.
The data statistics of Fullwiki+CNA setting are listed in Table \ref{tab:stat}.
This is a different evaluation from the original HotpotQA fullwiki setting.
If any of the gold SFs are not included in the passage, the gold answer is replaced with a CNA label.

The implementations of the IRC and one-stage model followed the algorithms described in Section~\ref{ssec:model} and~\ref{ssec:imple}. Finally, if the CNA score was higher than a hyperparameter $\beta$, the model outputted the CNA label. 
$\beta$, was determined from 0 to 0.9 by 0.1 to maximize the answer’s F1 score for each model.

\subsubsection{Results and Discussion}

\paragraph{\textbf{Does the IRC model improve answer performance considering unanswerability?}}
As shown in Table \ref{tab:fullwiki}, the IRC model predicted the answer more accurately than the one-stage model did. We consider that the IRC model effectively recognized the irrelevance of the rationale to the query in the answer module. This is because that the IRC model determines the CNA prediction against the rationale for the sake of the consistency. 
In contrast, the one-stage model was not good at CNA detection because its input always had information that was unnecessary for answering the query.

The IRC model extracts the rationale by focusing on recall because the rationale must cover the text necessary for answering the query. Under-extraction of SFs with the correct answer, which results in a high precision score, causes the reasoning shortcut problem.
We can explain the low EM score of the IRC model similarly. 
If the model aims for a high EM score, it is inevitable that under-extraction will occur as much as over-extraction. This is not an acceptable situation for interpretable and reliable RC.


The IRC model outperformed the ablated models. This indicated that our end-to-end training enabled the rationale module to consider ease of reasoning.

\begin{table}[t]
	\begin{center}
		\caption{
		Performance of our models in the Fullwiki+CNA setting. }
		\scalebox{0.85}{
			\begin{tabular}{r|c|c|c|c|c|c}\hline
				& \multicolumn{2}{c|}{Answer}
				& \multicolumn{4}{c}{SFs}
				\\ \hline
				& EM & F1 & EM & Precision & Recall & F1 \\ \hline
				One-Stage Model & 75.7 & 79.2 & \textbf{33.5}  & \textbf{53.2} & 56.1 & \textbf{52.7} \\ \hline
				IRC model & \textbf{80.2} & \textbf{83.3} & 12.5 & 36.1 & 71.4 & 43.2 \\ \hline
				-- End2End Training & 66.6 & 69.9 & 8.43 & 35.4 & \textbf{76.9} & 45.1 \\
				-- $L^{\textrm{R}}$ in End2End Training & 74.6 & 77.0 & 0.662 & 21.2 & 58.7 & 29.8 \\
				-- $L^{\textrm{NA}}$ in End2End Training & 73.5 & 76.8 & 10.9  & 38.1 & 75.6 & 47.1\\
				\hline
		\end{tabular}}
		\label{tab:fullwiki}
		\end{center}
\end{table}
    
\paragraph{\textbf{Can the IRC model avoid the reasoning shortcut?}}
Then, we discuss the performance in terms of the CNA detection task. Table \ref{tab:cna} shows the results of the evaluation of whether the answer is CNA or not. 
Each example is positive if the answer is CNA.
The high performance in this evaluation suggests that the model avoids the reasoning shortcut, where the model answers by force from insufficient text.

The IRC model outperformed the one-stage model. This is the same tendency as in Table \ref{tab:fullwiki}. Both the IRC and one-stage models detected the CNA with an EM score of more than 85.6\%. This indicates that a label classification including CNA and negative sampling of the paragraphs are effective at avoiding the reasoning shortcut. 

We consider that the consistency between the SFs and answer prediction contributed to the CNA detection performance of the IRC model.
In comparison, the one-stage model that ignores consistency may not detect that the query is unanswerable because it predicts the answer without recognizing that it may have missed the necessary information in the passage.

\begin{table}[t]
	\begin{center}
		\caption{
		Performance of our models as the CNA detection model.}
		\begin{tabular}{r|c|c|c|c}\hline
				& Acc. & Precision & Recall & F1 \\ \hline
				One-Stage Model & 85.6 & \textbf{94.2} & 85.4 & 89.6\\ \hline
				IRC model & \textbf{88.7} & 90.5 & \textbf{94.3} & \textbf{92.4} \\
    			-- End2End Training & 77.9 & 90.9  & 77.1 & 83.5 \\		-- $L^{\textrm{R}}$ in End2End Training & 83.2 & 87.8 & 89.2 & 88.5 \\
				-- $L^{\textrm{NA}}$ in End2End Training & 83.5 & 93.9 & 82.6 & 87.9 \\\hline
		\end{tabular}
		\label{tab:cna}
		\end{center}
\end{table}

\paragraph{\textbf{In what examples do the models detect unanswerable queries?}}
We classified the queries with respect to the number of absent gold SFs in the passage.
Table \ref{tab:stat_cna} lists the CNA prediction ratio for each class. 
Lower CNA prediction ratio is better if all the gold SFs are extracted. Otherwise, higher is better.
2089 examples included all the gold SFs, so the model should predict
the gold answer in the examples. The others had insufficient information. The IRC model predicted CNA correctly for more than 92.8\% of the insufficient passages.
The one-stage model performed worse in the examples with one absent sentence. This is the most typical case of the reasoning shortcut, where the model outputs the answer from one sentence including the answer while ignoring the sentence linking the query to the answer sentence.

\begin{table}[t]
	\begin{center}
		\caption{
		CNA prediction ratio for each class.}
			\begin{tabular}{c|c|c|c}\hline
				\# Absent Gold SFs & \# Examples & IRC & One-Stage
                 \\\hline
                 0 ($\downarrow$) & 2089 & 25.3 & \textbf{13.3} \\
                 1 ($\uparrow$) & 3418 & \textbf{92.8} & 80.0 \\
                 2 ($\uparrow$) & 1504 & \textbf{96.8} & 94.8 \\
                 3+ ($\uparrow$) & 374 & \textbf{98.4} & 96.8 \\ \hline
		\end{tabular}
        \label{tab:stat_cna}
	\end{center}
\end{table}

We further classified the 2089 examples according to whether the predicted SFs are sufficient or not (i.e., whether the model predicted the SFs including all the gold SFs or not). Table \ref{tab:zero} shows the results. 
We found that, in 837 examples, the extraction module of the IRC model failed to extract gold SFs, and one-stage model had more failed extractions. In such insufficient cases, the answer module of the IRC model outperformed the one-stage model in the CNA prediction performance by 16.3\% (higher is better).
In sufficient cases, the one-stage model outperformed the answer module of the IRC model by 10.3 \% (lower is better). We observed that the answer module predicted the CNA label conservatively in addition to the conservative prediction of the extraction module. This is important for reliability and interpretability.

\begin{table}[t]
	\begin{center}
		\caption{
		CNA prediction ratio in examples including all gold SFs. }
			\begin{tabular}{c|c|c|c|c}\hline
				Is Sufficient & \multicolumn{2}{c|}{IRC} & \multicolumn{2}{c}{One-Stage} \\
				& \# Examples & CNA & \# Examples & CNA
                 \\\hline
                 \checkmark & \textbf{1252} & 14.1 & 1178 & \textbf{3.82} \\
                  & \textbf{837} & \textbf{41.9} & 911 & 25.6 \\
		\end{tabular}
		\label{tab:zero}
	\end{center}
\end{table}

\subsection{Evaluation in HotpotQA without considering 'Can Not Answer'}
We compared the IRC and one-stage model with the previous models in the distractor setting.

\label{ssec:dist}
\subsubsection{Experimental Setup}  
The answer labels were $\{\textrm{`Yes', `No', `Span'}\}$. For training, we used the CNA label. Except for CNA, we regarded the label with the highest score to be the predicted label.

\begin{table}[t]
	\begin{center}
		\caption{
		Performance of models on the HotpotQA distractor setting leaderboard$^3$}
			\begin{tabular}{r|c|c|c|c}\hline
				& \multicolumn{2}{c|}{Answer}
				& \multicolumn{2}{c}{SFs}	
				\\ \hline
				& EM & F1 & EM & F1 \\ \hline
				Baseline \cite{hotpot} & 45.6 & 59.0 & 20.3 & 64.5 \\ 
				KGNN \cite{kgnn} & 50.8 & 65.8 & 38.7 & 76.8 \\
				QFE \cite{qfe} & 53.9 & 68.1 & 57.8 & 84.5 \\
				DFGN (base) \cite{dfgn} & 56.3 & 69.7 & 51.5 & 81.6 \\
				SAE (base) \cite{sae} & 60.4 & 73.6 & 56.9 & 84.6 \\ \hline
				HGN (large) \cite{hgn} & 66.1 & 79.4 & 60.3 & 87.3 \\ 
				SAE (large) \cite{sae} & 66.9 & 79.6 & 61.5 & 86.9 \\
				\hspace{6em} C2F Reader (large) \cite{c2f} & 68.0 & 81.3 & 60.8 & 87.6 \\
				\hline
				IRC model (base) & 58.6 & 72.5 & 36.7 & 79.4
                \\\hline
        \multicolumn{5}{l}{$^3$The unpublished models are not listed. The IRC model used BERT$_\textrm{base}$.}
		\end{tabular}
		\label{tab:test}
		\end{center}
\end{table}

\begin{table}[t]
	\begin{center}
		\caption{
		Performance of the models on the development set. }
			\begin{tabular}{r|c|c|c|c|c|c}\hline
				& \multicolumn{2}{c|}{Answer}
				& \multicolumn{4}{c}{SFs}
				\\ \hline
				& EM & F1 & EM & Precision & Recall & F1\\ \hline
				One-Stage Model & 58.4 & 72.8 & \textbf{54.3} & \textbf{87.3} & 85.7 & \textbf{85.0} \\ \hline
				IRC model & \textbf{58.6} & \textbf{72.9} & 36.6 & 76.0 & \textbf{89.1} & 79.8  \\ \hline
		\end{tabular}
		\label{tab:ablation}
		\end{center}
\end{table}

\begin{table*}[t]
	\begin{center}
	\caption{Outputs of our model with sufficient rationale.}
		\begin{tabular}{ccl}\hline
				\multicolumn{3}{l}{Ex.1 \quad $Q$: What was the sequel of the game that e was published by U.S. Gold in 1992?
				\quad $A$: Fade to Black } \\ \hline
				ground-truth & extracted & text \\ \hline
				\checkmark & \checkmark & 
				\parbox{38em}{\strut{} "Fade to Black (video game)" Fade to Black is an action-adventure game ... published by Electronic Arts.
		        \strut} \\ 
				& \checkmark & 
				\parbox{38em}{\strut{} "Fade to Black (video game)" It is the sequel to the 1992 video game `Flashback'.
		        \strut} \\ 
		         \checkmark & \checkmark & 
				\parbox{38em}{\strut{} "Flashback (1992 video game)" Flashback ... is a 1992 science fiction cinematic platform game ... published by U.S. Gold ...
		        \strut} \\ \hline\hline
				\multicolumn{3}{l}{Ex.2 \quad $Q$: Besides dísir, what is another Nordic term for a ghost?
				\quad $A$: Idisi } \\ \hline
				ground-truth & extracted & text \\ \hline
				\checkmark & \checkmark & 
				\parbox{38em}{\strut{} "North Germanic languages" ... `Nordic languages', a direct translation of the most common term used among Danish, Swedish and Norwegian scholars and laypeople.
		        \strut} \\ 
				 & \checkmark &
				\parbox{38em}{\strut{} "Dís" In Norse mythology, a dís (`lady', plural dísir) is a ghost, spirit or deity associated ...
		        \strut} \\ 
		        \checkmark & \checkmark & 
				\parbox{38em}{\strut{}"Dís" The North Germanic dísir and West Germanic Idisi are believed by some scholars to be related due to linguistic and mythological similarities ...
		        \strut} \\ \hline
		\end{tabular}
	\label{tab:qualitative}
	\end{center}
\end{table*}

\begin{table*}[t]
	\begin{center}
	\caption{Outputs of our model with insufficient rationale.}
		\begin{tabular}{ccl}\hline
				\multicolumn{3}{l}{\parbox{48em}{\strut{}Ex.1 \quad $Q$: Multiple award-winning actor Gary Sinise appeared in The Stand in 1994 - a miniseries based on a novel and screenplay by which noted author?
				\quad $A$: Stephen King \strut}} \\ \hline
				ground-truth & extracted  & text \\ \hline
				\checkmark & \checkmark &
				\parbox{38em}{\strut{}"The Stand (miniseries)" The Stand is a 1994 American television miniseries based on the novel of the same name by Stephen King.
		        \strut} \\ 
				 & \checkmark &
				\parbox{38em}{\strut{} "Gary Sinise" Gary Alan Sinise (...) is an American actor, director, and musician.
		        \strut} \\
				 \checkmark & &
				\parbox{38em}{\strut{} "Gary Sinise" Among other awards, he has won an Emmy Award, a Golden Globe Award, a star on Hollywood Walk of Fame and has been nominated for an Academy Award.
		        \strut} 
		        \\ \hline
		        \end{tabular}
	\label{tab:qualitative_insufficient}
	\end{center}
\end{table*}

\subsubsection{Results and Discussion}
\paragraph{\textbf{Does the IRC model improve the performance even when ignoring the unanswerability?}}
Table \ref{tab:test} shows the test results. 
The previous models have sophisticated structures, such as graph neural networks using entity linkage. 
However, the IRC model achieved comparable results to these models. In particular, our simple pipeline model consisting of a BERT layer and a linear layer had an F1 score that was 2.8 points higher than that of DFGN and 1.2 points lower than that of SAE, where both DFGN and SAE consisted of the BERT base model and the graph neural networks.


Table \ref{tab:ablation} shows the results of the evaluation on the development set. The IRC model performed comparably to the one-stage model. We consider that the IRC model has an effect on unanswerability detection, but has little effect on locating the answer. In addition, the IRC model could justify the model prediction. This shows that the IRC model can maintain prediction performance while providing a higher level of interpretability.


Similarly to the Fullwiki+CNA setting, the IRC model outperformed the one-stage model in terms of recall. The IRC model focuses on the recall score for interpretability. By contrast, the one-stage model extracts sentences with high precision scores. We consider that this is due to the difficulty in judging the irrelevance of the sentence to the query because of the relations among the sentences in the same paragraph. The CNA output seems to be useful to determine requirements conservatively.

\paragraph{\textbf{Qualitative Analysis of Extracted Rationale in case of sufficient extraction}}
The high recall prediction of the IRC model means that it redundantly extracts rationales.
However, we found that the predicted rationale often included a sentence useful for the answer prediction in addition to the ground-truth SFs. Table \ref{tab:qualitative} shows some examples. Double quotations represent the title of the article.

The first example is a case where the manual annotation is insufficient. The reasoning is incomplete without the second sentence. The second example is a case where we feel that the ground-truth SFs are sufficient for reasoning because we know `d\'isir' and `Idisi' are synonyms from the third sentence. However, we consider that the second sentence mentioning `ghost' contributes to the reasoning of the model. 
From another point of view, the infrequent words, `d\'isir' and `Idisi', are not well represented in the embedding space, so the second sentence plays the role of prior knowledge. 

\paragraph{\textbf{Qualitative Analysis of Extracted Rationale in case of insufficient extraction}}
Though the IRC model extracts rationales conservatively, there are some examples where the model extracts insufficient sentences but predicts a non-CNA answer. 
However, we observed that such extractions also have enough information for reasoning.

\cite{shortcut2} classifies the examples in the distractor setting in four categories: multi-hop, weak-distractors, redundant evidence, and non-compositional single-hop. 
Redundant evidence means that the query has a redundant entity. We found that sentence not included in the gold SFs usually serve as substitutes for an absent gold SF or had the effect of increasing the confidence of the answer.

Table \ref{tab:qualitative_insufficient} shows an example of redundant evidence examples with the redundant entity of Gary Sinese because the query is valid without mentioning him. The only one gold SF is sufficient for reasoning. However, the non-gold sentence gave basic knowledge about Gary Sinese. We consider that the supplementary information has an effect on confidence in the model even if the information is on a redundant entity.

\section{Conclusion}
We defined interpretable reading comprehension (IRC) as the ability of a model to justify an answer behind valid reasoning based on sufficient information. It is implemented as a pipelined model with a `Can Not Answer' label. 
The IRC model regards the SFs as information passed between the modules in order to provide the actual rationale of the answer. The CNA output improves the reliability of the answer prediction by detecting reasoning with insufficient rationale. 
We believe that our study advances the interpretability of RC to the next level and the IRC model resolves the issues of reading comprehension, such as fact checking.


\bibliography{theme}
\bibliographystyle{IEEEtran}

\end{document}